\documentclass[preprint,12pt]{Styles/elsarticle}

\PassOptionsToPackage{hyphens}{url}
\usepackage{hyperref}
\usepackage{float}
\usepackage{verbatim} 
\restylefloat{figure}
\restylefloat{table}
\usepackage[utf8]{inputenc} 
\usepackage[T1]{fontenc}    
\usepackage{url}            
\usepackage{booktabs}       
\usepackage{amsfonts}       
\usepackage{nicefrac}       
\usepackage{microtype}      
\usepackage{xcolor}         
\usepackage{amsmath}
\usepackage{mathtools}
\usepackage{amssymb}
\usepackage{mathdesign}
\usepackage{algorithm}
\usepackage{algorithmic}
\algsetup{linenosize=\tiny}
\usepackage{dsfont}
\usepackage{graphicx}
\usepackage{array, booktabs, makecell}
\usepackage{siunitx}
\DeclareMathOperator*{\argmin}{arg\,min}
\usepackage{amssymb} 
\usepackage{bbm}
\usepackage{amsthm}
\usepackage{caption}
\usepackage{subcaption}
\theoremstyle{definition}
\newtheorem{exmp}{Example}

\usepackage{glossaries}
\makeglossaries
\newacronym{LHS}{LHS}{left hand side}
\newacronym{RHS}{RHS}{right hand side}

\newcommand{\Danit}[2]{\textcolor{purple}{*** DSA: #1
***}}

\journal{Engineering Applications of Artificial Intelligence}
\usepackage{lipsum}
\makeatletter
\def\ps@pprintTitle{%
 \let\@oddhead\@empty
 \let\@evenhead\@empty
 \def\@oddfoot{}%
 \let\@evenfoot\@oddfoot}
\makeatother
\begin{document}
\begin{frontmatter}
\title{Adaptive Learning for the Resource-Constrained Classification Problem}


\author[inst1]{Danit Shifman Abukasis}
\author[inst1]{Izack Cohen}
\author[inst2]{Xiaochen Xian}
\author[inst3]{Kejun Huang}
\author[inst1]{Gonen Singer}

\affiliation[inst1]{organization={Faculty of Engineering, Bar-Ilan University}, city={Ramat Gan},addressline={5290002}, country={Israel}}
\affiliation[inst2]{organization={Industrial and Systems Engineering Department, University of Florida},city={Gainesville}, addressline={FL 32611}, country={USA}}
\affiliation[inst3]{organization={Department of Computer and Information Science and Engineering, University of Florida},city={Gainesville}, addressline={FL 32611}, country={USA}}

\begin{abstract}

Resource-constrained classification tasks are common in real-world applications such as allocating tests for disease diagnosis, hiring decisions when filling a limited number of positions, and defect detection in manufacturing settings under a limited inspection budget. Typical classification algorithms treat the learning process and the resource constraints as two separate and sequential tasks. Here we design an adaptive learning approach that considers resource constraints and learning jointly by iteratively fine-tuning misclassification costs. Via a structured experimental study using a publicly available data set, we evaluate a decision tree classifier that utilizes the proposed approach. The adaptive learning approach performs significantly better than alternative approaches, especially for difficult classification problems in which the performance of common approaches may be unsatisfactory. We envision the adaptive learning approach as an important addition to the repertoire of techniques for handling resource-constrained classification problems.
\end{abstract}


\begin{keyword}
classification; resource constraints; resource allocation; cost-sensitive learning; adaptive learning 



\end{keyword}

\end{frontmatter}



\section{Introduction}
\label{sec:introduction}

Classification, one of the most popular machine learning tasks, is frequently used for decision-making across a variety of real-world applications such as classifying the severity of diseases \cite{nabi2019characterization}, scheduling and allocating resources \cite{pessach2020employees}, and defect classification for manufacturing processes \cite{yang2009online, anzanello2012multicriteria,marques2018improving, qiu2020process}. Classification applications are typically associated with misclassification costs and benefits as a result of incorrect and correct classification, respectively. Many studies have focused on cost-sensitive classification approaches \cite{elkan2001foundations,ling2008cost-sensitive,rokach2008pessimistic,cui2019class-balanced,li2018cost-sensitive,zhang2020cost-sensitive} in an effort to reduce the costs of misclassification. We illustrate the concept of imbalanced misclassification costs using the current and real-world example of classifying COVID-19 patients. Incorrectly classifying an ill patient as healthy may put this patient's life at risk as well as others by allowing the ill person to circulate among healthy persons and infect them (an intangible cost, usually determined by the judicial system). Classifying a healthy individual as a COVID-19 patient, on the other hand, may lead to unnecessary treatment, misuse of medical resources and cause unnecessary financial hardship to the individual and the general economy. Many studies have applied cost-sensitive approaches to handling imbalanced classification problems \cite{dong2020cost,buda2018systematic} where the decision maker is interested in detecting the positive cases. 

There are four main approaches for making a classifier cost-sensitive: (i) changing the distribution of classes using over- and under-sampling within the training data set (i.e., preprocessing of the training data) to reduce misclassification costs \cite{elkan2001foundations,ling2008cost-sensitive}, denoted hereafter approach A1; (ii) changing the data set according to the misclassified samples of the cost-insensitive classifiers and their error costs (post-processing the training data) using a boosting approach in ensemble learning methods \cite{zhang2020cost-sensitive,li2005cost}, denoted hereafter approach A2; (iii) incorporating meta-learning methods on outputs of cost-insensitive learners using threshold driven techniques in favor of utilizing the probability estimations for the classes \cite{elkan2001foundations,ling2008cost-sensitive,hernandez-orallo2012unified,domingos1999metacost}, hereafter denoted  A3; (iv) directly incorporating cost-sensitive capabilities into a learning algorithm, i.e., an algorithm-level solution that adapts existing learning methods so they are biased towards classes with high misclassification costs, usually presented by minority classes \cite{ling2008cost-sensitive,zadrozny2001obtaining}. We denote this approach as A4.

To summarize, A1 and A2 are data-level methods that change classes' distributions or their sampling weights. Since these approaches change the training data set, they can cause the classifier to be biased towards oversampled data or against the undersampled data and, therefore, may distort the interpretability of the generated classifying models. Furthermore, these methods have little effect on the classifiers, as claimed by Elkan \cite{elkan2001foundations}. A3, which does not change the training data, classifies instances based on the probability estimation output of a cost-insensitive classifier. As A1 and A2, it does not directly affect the construction of the classifier. The performance of A3, moreover, may lead to unreliable results when it relies on decision trees and Bayesian networks that tend to provide inaccurate probability estimations \cite{zadrozny2001obtaining,domingos1996beyond,provost2000well-trained,zhao2008instance}. A4 was used in previous studies for constructing decision tree models that take the misclassification costs into account when selecting the best node split and for tree pruning  \cite{elkan2001foundations,li2018cost-sensitive,ting2002instance-weighting,rokach2009classification,chaabane2019enhancing,bahnsen2015example}. Some studies incorporated the costs associated with each chosen attribute in the path traced from the tree node to the leaf node in the construction of the decision tree models \cite{zhang2005missing,ling2006test,yang2006test-cost,zhao2017cost}, while others incorporated constraints reflecting limited budgets when choosing attributes \cite{qin2004cost,chen2016time-constrained}. Despite the focus of these studies on cost-sensitive classifiers, none of them considers a constraint on the number of instances classified into specific classes due to scarce resources. Moreover, most cost-sensitive studies aim to solve classification problems with known misclassification costs or use the cost-sensitive approaches to adjust insensitive classifiers to handle imbalanced data. The latter studies use the misclassification costs to reflect the imbalance of the classes in the training data set. With this lacunae in mind, our study, the first to do so, utilizes cost-sensitive learning approaches to incorporate resource constraints into the learning process.   

To understand our motivation we return to our COVID-19 example. Consider diagnosing COVID-19 patients using a limited number of inspections. How should the test kits be allocated among people who are suspected as being ill? One allocation can be to prioritize those who were exposed to more people (a specific class) in favor of preventing infections. Yet, another reasonable decision is to prioritize the identification of people who are at a higher risk of developing a serious disease (another class) to increase their chances of recovery.

The typical approach for classification under resource constraints is to feed the classification results as exogenous parameters into an optimization model that finds (near-) optimal resource allocation \cite{pessach2020employees,sarkar2015improving}. Such an approach usually requires formulating a resource allocation optimization problem that gets the
classification results as input parameters and the limited resources are the constraints \cite{pessach2020employees,li2018cost-sensitive,zhang2020cost-sensitive,sarkar2015improving}. In these approaches a classification model, which is built based on a labeled training data set, is applied to a new data set that is independent of the training data, which we refer to as the test data. Constraints are addressed via an optimization
model. Few studies blend the resource constraints by post-processing the training data via varying the weights of instances (A2) \cite{israeli2019constraint}. The limitation of these approaches is that the classification model does not consider the constraints during its learning process.


The main contributions of this research are:
\begin{enumerate}
    \item Methodologically, this is the first research to develop an integrated framework that incorporates resource constraints into the learning process to solve resource constrained-based classification models. For this, we use cost-sensitive approaches A3 and A4, without changing the class distributions and distorting the interpretability of the generated classifying models. 
\item From a modeling perspective, the suggested approach can
handle settings in which resource constraints dynamically change over time. 
For example, consider a COVID-19 test kit purchase order that was delayed, and thus the number of the available test kits is smaller than anticipated. As a consequence, the classifier model should be updated according to the updated resource constraint -- which ability our model has. Another unique feature of the suggested modeling approach is that it does not require exact knowledge about the misclassification costs, which are typically hard to obtain. In other words, the algorithm is assured to converge to the optimal solution even when starting from arbitrary cost values. 
\item Testing the suggested approach demonstrates its favorable performance compared to approaches that do not consider learning and resource constraints together.
\end{enumerate}


The rest of the paper is organized as follows. Section \ref{sec:perliminaries} introduces a motivation example and the mathematical background on cost-sensitive approaches. In Section \ref{sec:approach}, we introduce our proposed approach for handling misclassification cost problems with resource constraints and describe the adaptive cost-sensitive learning algorithm for resource-constrained classification problems. The numerical experiments and results are presented in Section \ref{sec:experiment}. Finally, conclusions are presented in Section \ref{sec:conclusion}. 

\section{Preliminaries and Mathematical Background}
\label{sec:perliminaries}
This section presents the necessary notation and background as well as common classification approaches before we develop, in Section \ref{sec:approach}, an adaptive cost-sensitive and resource-constrained learning approach.

\subsection{Running Example for a Constrained Cost-Sensitive Classification Problem} \label{sec:motivation example}

We use the following running example throughout the paper to illustrate the suggested approach. Assume we have $100$ COVID-19 test kits and $1,000$ people arrived at our hospital and are waiting to be tested. Obviously, the test kits can be allocated only to $10\%$ of the people. Based on an initial assessment of these people upon their arrival, it turns out that 8$\%$ of them lost their sense of taste (feature $f_1$) and 20$\%$ had a high fever (feature $f_2$) -- both of which are classic COVID-19 symptoms. All the necessary information about the people is known including the fact that $100$ of them were actually infected with COVID-19 (i.e., this is a supervised training data set). The cost of incorrectly classifying a COVID-19 patient as healthy (false negative) is 19 times larger ($\$19$) than classifying a healthy person as infected (false positive, $\$1$). The objective is to assign the available test kits to the infected patients such that the misclassification costs are minimized.

\begin{figure}[!h]
	\centerline{\includegraphics[scale=0.5]{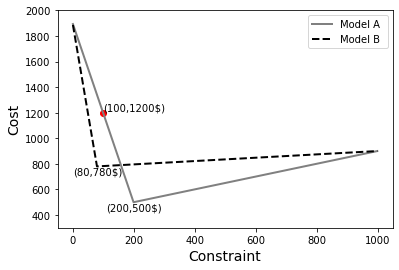}}
	\caption{Total misclassification costs as a function of the number of available COVID-19 test kits}
	\label{fig:motivation_example}
\end{figure}

Figure \ref{fig:motivation_example} shows the misclassification costs when trying to diagnose COVID-19 patients among the $1,000$ people as a function of the number of COVID-19 test kits available for the following two classification models.

Model A illustrates the case of classifying the $1000$ people according to a conventional \textit{unconstrained} cost-sensitive decision tree model based on the training data collected upon their arrival at the hospital. The model classifies the people based on feature $f_2$ (high fever). According to this model, $200$ people are classified as positive, of which $80$ are true positive and $120$ are negative (false positive), yielding a precision of 40\%; there are  $20$ false negatives. As can be seen, the minimal total cost is $120\cdot \$1 +20 \cdot \$19 = \$500$. Since there are only $100$ test kits available, only 100 people can be classified as positive. A naive decision-making approach would allocate the tests to the $100$ people with the highest probability of being positive from the 200 positive classifications. Since all $200$ people have the same probability according to the classification model, $100$ of them would be randomly selected to be tested. 

Of these $100$ people, we expect $80\cdot0.5 = 40$ to be true positive and $60$ false positive based on the $40\%$ precision, and the expected number of false negatives would increase to $60$. As a result, the expected cost for the constrained classification would be higher, $60 \cdot \$1 + 60 \cdot \$ 19 = \$1,200$, as shown by the red point in Figure \ref{fig:motivation_example}. 

Model B presents the proposed method that considers the resource constraint of $100$ tests and the features of the potentially COVID-19-infected people. According to this model, $80$ people are classified as positive, of which $20$ are false positive. Therefore, there are also $40$ false negatives. As a result, the total cost for $100$ COVID-19 tests is $20 \cdot \$1 + 40 \cdot \$ 19= \$ 780$, with a precision of 75\%.  Although Model B's minimum cost is higher than that of Model A for the unconstrained classification problem (780 compared to 500, respectively), for the considered constraint, Model B achieves a cost that is 35\% lower compared to Model A. We note that a decision-maker who wants to use all the 100 tests would still prefer Model B compared to Model A at a cost of \$$782.6$. Moreover, in binary classification problems, which are considered in this work, with maximum utilization of resources, a model that produces a lower cost will necessarily produce higher accuracy and precision and vice versa (see Section \ref{sec:constraint_line}).  

\subsection{Decision Threshold Approach for Misclassification Cost Problems}
In this section, we introduce basic concepts and definitions related to the binary cost-sensitive learning problem.

We denote a fully labeled training data set with $N$ instances and $k$ features as $D=\{(x_i,y_i )|i=1,\dots,N\}$, where $x_i=(v_{i1},v_{i2},\dots,v_{ik} )\subseteq R^{k}, i=1,\dots,N$\ denotes a vector of features, for instance $i$, and $y_i\in\{0,1\}$ is the value of the dependent variable (denoted hereafter as \textit{label}) where $0$ and $1$ are the negative and positive classes, respectively. 
$\mathcal{M}$ is a classification model (classifier) that maps $x_i$ to its predicted label $y_i$. A cost-sensitive classifier aims to minimize the misclassification costs over a considered data set.

Following \cite{elkan2001foundations}, we denote $\mathcal{C}$ as the misclassification cost matrix where the cost of labeling instance $x$ as class $l$, when the actual class is $j$, is $c(l,j)\geqslant 0$  $\forall l,j\in \{0,1\}$. We assume that $c(l,j)=0, \forall l=j$ following \citeauthor{ling2008cost-sensitive} \cite{ling2008cost-sensitive} who showed that any cost matrix can be transformed into an equivalent matrix in which correct classifications have a zero cost and with the same classification decisions. Table \ref{tbl:cost_matrix} presents a cost matrix $\mathcal{C}$ for the binary case, which is where we put our focus. $c(0,0)$ and $c(1,1)$ represent the costs of true negative ($TN$) and true positive ($TP$) classifications, respectively, and $c(1,0),c(0,1)$ represent the costs of false positive ($FP$) and false negative ($FN$) classifications, respectively. In many real-world problems, the cost of an $FN$ classification is higher than the cost of an $FP$, i.e., $c(0,1)>c(1,0)$ as reflected in the example in Section \ref{sec:motivation example}.    
\begin{table}[!t]
	\renewcommand{\arraystretch}{1.2}
	\caption{Binary cost matrix}
	\label{tbl:cost_matrix}
	\begin{tabular}{ l | c| c }
		& \textbf{Actual 0} & \textbf{Actual 1} \\ \hline
		\textbf{Predict 0} & $c(0,0)$ & $c(0,1)$ \\ \hline
		\textbf{Predict 1} & $c(1,0)$ & $c(1,1)$ \\ \hline
	\end{tabular}
	\centering
\end{table}
The expected loss (or in our case, the cost) of labeling instance $x$ as class $l$ is
\begin{equation}\label{eq:expected_loss}
	L(x_i,l)=\sum_{j=0}^{1} P(j|x_i)c(l,j),
\end{equation}
where $P(j|x_i)$ is the probability that the true label of instance $x_i$ is $j$, obtained from a classification model $\mathcal{M}$. Therefore, the classification that minimizes the expected loss is
\begin{equation}\label{eq:argmin_loss}
	\hat{y_i}=\argmin _{l\in \{0,1\}}L(x_i,l).
\end{equation}
As mentioned in the Introduction, a common approach for solving problems with misclassification costs involves using a threshold on the probability outputs to minimize the misclassification costs (denoted as A3). In general, a classification model $\mathcal{M}$ computes the probability that the true label of instance $x_i$ is the positive class -- that is, $ s_i = P(1|x_i)$. The classification is made by comparing $s_i$ to some threshold $\tau$. If $s_i \geqslant \tau$, then $\hat{y_i}=1$; that is, instance $i$ is given the positive class label. Otherwise, it is labeled as negative, $\hat{y_i}=0$.

\citeauthor{elkan2001foundations} \cite{elkan2001foundations} proved that the optimal threshold $\tau$ for a binary cost-sensitive classifier is 
\begin{equation}\label{eq:binary_threshold}
	\tau=\dfrac{c(1,0)-c(0,0)}{c(0,1)+c(1,0)-c(0,0)-c(1,1)}. 
\end{equation}
If $c(0,0)=c(1,1)=0$, then the threshold is reduced to
\begin{equation}\label{eq:reduced_threshold}
	\tau=\dfrac{c(1,0)}{c(0,1)+c(1,0)}.
\end{equation}
Therefore, if $s_i\geqslant \tau$, a positive classification is optimal. Note that the threshold $\tau$, which is determined exclusively by $\mathcal{C}$, is static and independent of $D$. 

Finally, the classification follows
\begin{equation}\label{eq:classification_step_function}
    \hat{y_i} = \theta(s_i - \tau),
\end{equation}
where $ \theta(z) $ is a step function such that
\begin{equation}\label{eq:theta_indicator}
    \theta(z) =
     \begin{cases}
     	1, & \text{if}\ z \geqslant 0 \\
     	0, & \text{otherwise}.
     \end{cases}
\end{equation}

We use the idea of a threshold to accommodate the resource constraint into the cost-sensitive learning process.

\subsection{Iso-Loss Line}

\begin{figure}[!t]
	\centerline{\includegraphics[scale=0.5]{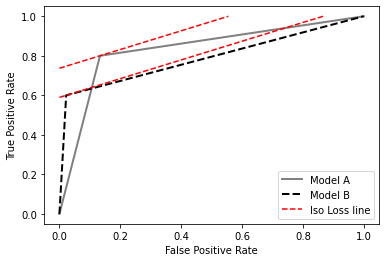}}
	\caption{Iso-Loss line and ROC curve for COVID-19 motivation example}
	\label{fig:iso_line}
\end{figure}

This section introduces the iso-performance line introduced by \citeauthor{provost2001robust} \cite{provost2001robust}. The iso-performance line connects the loss to the receiver operating characteristic (ROC) curve and enables the best classifier to be chosen in terms of loss.
The ROC space is defined by the y-axis' true positive rate (tpr), also known as recall, which is calculated as the number of true positive classifications divided by the number of all actual positives, and the x-axis' false positive rate (fpr), also known as fall-out, which is the number of negatives wrongly classified as positives divided by the number of all actual negatives.

Following \cite{flach2004many}, the loss of a classifier over a data set can be defined as
\begin{gather}\label{eq:loss_iso_with_prob}
    	\mathcal{L} = \sum_{i=1}^N \sum_{l\in(0,1)} L(x_i, l) = \\ c(0,1) (1-tpr)|D^p| 
    	+ c(1,0)  fpr|D^n|,  
    	\nonumber
\end{gather} 
where $|D^p|$ and $|D^n|$ are the number of total actual positive and negative instances in data set $D$, respectively. One can find the optimal point $(fpr^*,tpr^*)$, among all possible points in the ROC graph, that achieves the minimum costs by,

\begin{equation}\label{eq:argmin_loss_ROC}
	(fpr^*,tpr^*)=\argmin _{(\forall {fpr},{tpr})}\mathcal{L}.
\end{equation}

Rearranging Eq. (\ref{eq:loss_iso_with_prob}), we obtain
\begin{equation}\label{eq:iso_with_prob}
    tpr = \frac{c(1,0) |D^n|}{c(0,1)  |D^p|} fpr - (\frac{\mathcal{L}  }{c(0,1)  |D^p|} - 1),
\end{equation}
which is called the iso-loss line reflecting the relationship between losses $fpr$ and $tpr$. As can be seen from Eq. \ref{eq:iso_with_prob}, the slope of the line is constant and determined by the properties of the data set and the cost structure. It can also be seen that for a given $fpr$, a higher line represents a lower loss. Figure \ref{fig:iso_line} presents both the ROC curves for Models A and B from our running example in Section \ref{sec:motivation example} as well as the iso-loss lines. We use the ROC graph and iso-loss lines to present the direction to which we would like to change the ROC graph of a classifier to reduce the cost by influencing the learning process (oriented learning). In Section \ref{sec:constraint_line}, we add a constraint to the ROC graph, such that the limit of the number of instances that can be classified per class would be taken into account as part of the learning process. 

\subsection{Post-Classification Resource Constraints}
\label{sec:Resource_Constraint}
Our main assumption is that a budget constraint limits the number of instances that can be classified to a class. Accordingly, we define the maximal number of instances to a class $k$ to be  $\mathbb{B}(k)$. 

Without loss of generality for the binary case, we assume that the constraint is on the positive class, which is represented by the value $1$. Hence, the constraint that limits the number of per-class classifications can be defined by

\begin{equation}\label{eq:constraint_def}
    \sum_{i=1}^{N'} \hat{y_i} \leqslant \mathbb{B}(1),
\end{equation}
where $N'$ is the number of instances in the considered data set (typically, the test data set) and $\hat{y_i}\in\{0,1\}$ is the predicted output of instance $i$. If the optimal number of positive classifications, obtained from a cost-sensitive classifier $\mathcal{M}$, is larger than $\mathbb{B}(1)$, the resource constraint is violated. In such a case one needs to decrease the number of positive classifications to satisfy the constraint. This necessarily decreases the sum of true and false positives and increases the sum of true and false negatives. 

The common approach to combining the classification and resource constraint is to use the results of the classifier $\mathcal{M}$ as input parameters to an optimization problem \cite{pessach2020employees,sarkar2015improving}. 

The optimization problem can be formulated based on Eqs. (\ref{eq:argmin_loss}) and (\ref{eq:constraint_def}) as 
\begin{gather}\label{eq:optimization problem def}
    Min \sum_{i=1}^{N'} \sum_{j=0}^{1} P(j|x_i)c(\hat{y_i},j)\\
    s.t. \sum_{i=1}^{N'}\hat{y_i}\leqslant \mathbb{B}(1). \nonumber
\end{gather}
The solution approach for this problem 
classifies the instances with the highest probability of being positive as positives, which expresses a belief that these instances relate to the positive class. Without loss of generality, we assume that probability outputs, defined as  $s_i = P(1|x_i)$, are sorted in a decreasing order, i.e.,  $(s_1,s_2,\dots,s_i,s_j,\dots)$ such that $\forall j>i \rightarrow s_i\geqslant s_j$. Then, $x_j$ may be classified as positive only if 
\begin{equation}\label{eq:classification_by_constraint}
    \sum_{\forall s_i\geqslant s_j} \hat{y_i} < \mathbb B(1),
\end{equation}
meaning that the constraint is still  not satisfied. In other words, the number of positive classifications for instances with score $s_i\geqslant s_j$  is smaller than the value of the constraint, so there is room for more positive classifications.
According to the above-mentioned naive approach, the constraint is considered only during the decision-making phase. We will use this approach for benchmarking purposes, in the experimental study.

\section{Adaptive Cost-Sensitive Learning with Resource Constraints}
\label{sec:approach}
We now introduce the proposed approach for embedding resource constraints into the learning process.

The objective is to minimize the cost while utilizing all the resources (e.g., if the constrained resource is $100$ COVID-19 test kits, the decision maker should allocate all these kits). The idea is that the suggested adaptive approach combines knowledge about classification results of the training data set and the resource constraint to iteratively adjust the misclassification costs. The new misclassification costs are used to update a cost-sensitive classifier. The process repeats until the resource constraint is satisfied with the optimal threshold in Eq. (\ref{eq:reduced_threshold}).
 
\subsection{Constraint Line in a ROC Graph}
\label{sec:constraint_line}
Here we present the resource constraint within the ROC graph to define the feasible classification area inside it, and search for the optimal point ($fpr^*$, $tpr^*$) that yields the minimum cost, subject to the constraint.

The decision threshold choice, which is presented on the ROC graph of the training data set, projects the resource constraint of the test data set onto the training data set; that is, the ratio between the resource constraints of the testing and training data sets will be equal to the ratio between the sizes of the testing and training data sets. Since the maximal number of positive instances in the training data is denoted by $\mathcal{B}(1)$, the following inequality holds 
\begin{equation}\label{eq:constraint_for_roc}
	\mathcal{B}(1) \geqslant |D^n|\cdot fpr + |D^p| \cdot tpr,
\end{equation}
where the right hand-side of the equation is the sum of instances \textit{predicted} as positive.

Rearranging the inequality and extracting the line on which the number of positive predictions is forced to equal $	\mathcal{B}(1)$ (i.e., resources are fully utilized) leads to the following constraint line that defines the \textit{feasible area} of the ROC space:
\begin{equation}\label{eq:constraint_line}
	tpr = -\frac{|D^n|}{|D^p|}\cdot fpr + \frac{\mathcal{B}(1)}{|D^p|}.
\end{equation}

The constraint line intersects with every ROC curve of a classifier. By definition, at the intersection point, the number of positive classifications equals the resource constraint, i.e. $\sum_{i=1}^{N} \hat{y_i} = \mathcal{B}(1)$. Each point on the ROC curve, which is below the constraint line of Eq. (\ref{eq:constraint_line}), represents a solution in which the number of positive classifications is smaller than the resource constraint, $\sum_{i=1}^{N} \hat{y_i} < \mathcal{B}(1)$. The minimum cost solution of a classifier that satisfies the resource constraint is a point on the ROC curve within the feasible area that has a minimum loss as calculated by Eq. (\ref{eq:argmin_loss_ROC}).




The theoretical minimum cost is achieved for the case $|D^p| \geqslant \mathcal{B}(1)$, when all positive classifications $\mathcal{B}(1)$ are correct, i.e., $fpr=0$  or for the case of $|D^p|< \mathcal{B}(1)$, when the positive classifications $\mathcal{B}(1)$ capture all actual positives, i.e., $tpr=1$. Note that for both cases, we force complete resource utilization, which means that there are exactly $\mathcal{B}(1)$ positive classifications. The cost at these points is achieved using 
Eq. (\ref{eq:loss_iso_with_prob})

\begin{equation}\label{eq:minimum_cost}
	\mathcal{L} =  
	\begin{cases}
	    c(0,1)\cdot
	({|D^p|-\mathcal{B}(1)}), & \text{if}\ |D^p|\geqslant \mathcal{B}(1)\\
	    c(1,0)\cdot
	({\mathcal{B}(1)-|D^p|}), & \text{otherwise}
	\end{cases}. 
\end{equation}

\begin{figure}[!t]
	\centerline{\includegraphics[scale=0.5]{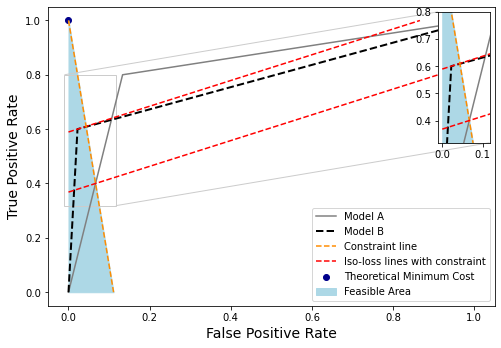}}
	\caption{Feasible area and optimal iso-lines for the COVID-19 running example}
	\label{fig:feasible_area_example_2}
\end{figure}

Figure \ref{fig:feasible_area_example_2} illustrates, for our running example, the feasible area defined by the constraint line and the optimal iso-loss lines of Models A and B within the feasible area. It can be seen that the optimal iso-loss line of Model A, which satisfies the resource constraint, is different from the optimal iso-loss line of unconstrained Model A in Figure \ref{fig:iso_line}, reflecting a much higher cost than the iso-loss line of Model B. Note that the optimal iso-line of Model B with minimum cost is achieved when the resource constraint is not fully utilized. The theoretical minimum cost of Eq. (\ref{eq:minimum_cost}) is presented as the intersection point with the tpr axis. In Section \ref{sec:dynamic_threshold}, we will use the feasible area and iso-loss lines to develop an adaptive cost-sensitive learning algorithm to solve resource-constrained classification problems, in which the resource constraint is fully utilized, i.e., $\sum_{i=1}^{N} \hat{y_i} = \mathcal{B}(1)$; such a situation is manifested in the intersection points of the ROC curves and the constraint line. Since a higher intersection point reflects a higher tpr with a higher iso-loss line, as presented in Eq. (\ref{eq:iso_with_prob}), the suggested algorithm aims to achieve a ROC curve with the highest possible intersection point. We also note that a higher intersection point represents a higher iso-accuracy line, which is defined by Eq. (\ref{eq:iso_with_prob}) with $c(0,1)=c(1,0)$. Therefore, an algorithm that improves the cost necessarily improves the accuracy. Since the resource constraint is fully utilized, the number of positive classifications ($TP+FP$) is constant. Moreover, since the number of actual negatives ($TN+FP$) is also constant, the expression ($TP-TN$) is constant as well. Thus, an algorithm, which improves the accuracy ($\frac{TP+TN}{TP+TN+FP+FN}$), necessarily improves the precision ($\frac{TP}{TP+FP}$). This occurs because the improvement in accuracy may be due to an increase in both $TP$ and $TN$ (by the same value needed to maintain a constant difference), which  causes a direct increase in precision (an increase in the value of the numerator, while maintaining the value of the denominator constant). To summarize, an algorithm that improves the cost of a resource-constrained classification problem, in which the resource constraint is fully utilized, necessarily improves both accuracy and precision and vice versa. 



\subsection{Classifier Dependent Threshold for Resource-Constrained Classification Problems} \label{sec:dynamic_threshold}
Assume we have a classifier that splits a data set $D$ into subsets, where a subset contains a group of instances that are predicted to belong to a specific class. This assumption holds for decision tree based algorithms, which identify subsets of instances in the training data set that share common patterns and that results in the same prediction probability. In other word, each instance in subset $D_k$ of $D$ has the same prediction probability $S_k$, i.e., $s_j=S_k,\forall x_j\in D_{k}\ $ such that $\bigcup\limits_{\forall k} D_k=D$ and the number of instances within a subset is $|D_k|$. 


We now determine, for a such classifier, a threshold that satisfies the resource constraint and yields the minimum cost. 


The probability scores of the subsets are sorted in a decreasing order into vector $(S_1,S_2,\dots,S_l,S_m,\dots)$ such that $\forall m>l \rightarrow S_l\geqslant S_m$. We denote $\tau_{d}$ as a classifier dependent threshold, which is calculated as follows:

\begin{equation}
\label{eq:adj_threshold}
\tau_{d} = \max\{\tau,\max_{k} \theta(S_k-\tau)\cdot\theta\bigg(\sum_{\forall S_m\geqslant S_k}|D_m|-\mathcal B(1)\bigg)\cdot S_k\},
\end{equation}
where the function $\theta(\cdot)$ was defined by Eq. \ref{eq:theta_indicator}. The result of the expression gets the highest $S_k$ value for which the total number of positive classifications is equal to or greater than $\mathcal B(1)$. 
The instances of a data set are classified based on the classifier dependent threshold. All instances within the subsets with $S_k>\tau_d$ are classified as positives and all instances within subsets for which $S_k<\tau_d$ are classified as negatives. The instances within the subset for which $S_k=\tau_d$ are allocated to the positives, according to the following proportion:

\begin{equation}
\label{eq:proportion_all}
p=\dfrac{\mathcal B(1)-\sum_{\{S_k>\tau_d:\forall k\}} |D^k|}{|D^k\mid S_k=\tau_d|}.
\end{equation}

The proportion is calculated as the number of allowed (remaining) positive classifications after some were allocated to the subsets with $S_k>\tau_d$ divided by the number of instances within the subset $k$ for which $S_k=\tau_d$.  
Algorithm \ref{alg:classification_based_dynamic_thershold} presents the classification mechanism of a data set, according to the classifier dependent threshold. For illustration, consider Example \ref{exp:dynamic_threshold}.
$$
$$
\begin{exmp}\label{exp:dynamic_threshold}
Let us say that we are given $(S_1,S_2,S_3,S_4)=$ $(0.8,0.7,0.6,0.5)$, $(|D_1|,|D_2|,|D_3|,|D_4|)=$ $(5,4,3,2)$, $\tau=0.55$ and $\mathcal B(1)=7$. Applying Eq. (\ref{eq:adj_threshold}) results in $\tau_d = S_2=0.7$. 

Under such circumstances, the classifier will allocate the scarce resource of positive classification as follows; the five instances within Subset 1 will be classified first as positives and the remaining two positives will be randomly allocated to two of the four instances within Subset 2. 
We note that for any $\mathcal B(1)\geqslant14$ (the unconstrained case), the threshold is $\tau_d=\tau=0.55$ since the right-hand term in the $\max\{\tau,\cdot\}$ of Eq. \ref{eq:adj_threshold} is 0. 

\end{exmp}

\begin{algorithm}[!t]
    \caption{Classification of the training data set based on the classifier dependent threshold}
            
    \begin{algorithmic}\label{alg:classification_based_dynamic_thershold}
        \STATE {\bfseries Input:} Cost dependent threshold $\tau$, training data sets ${D}=(X,Y)$, subsets of training data set $(D_1, D_2,\dots)$, ordered scores vector $S=(S_1, S_2,\dots)$, resource constraint $\mathcal{B}(1)$
        
        \STATE {\bfseries Output:} Classification results $\hat{Y}$ 
        \STATE {\bfseries Initialize:} $\tau_d \leftarrow$ use Eq. (\ref{eq:adj_threshold})
        
        \FOR{$S_k \in S$}
            \IF{$S_k > \tau_d $}
                \STATE $ \hat{y}_i \leftarrow 1,  \forall i: P(1|x_i) = S_k$
            \ELSIF{$S_k < \tau_d $}
                \STATE $ \hat{y}_i \leftarrow 0,  \forall i: P(1|x_i) = S_k$
            \ELSE 
                \STATE $p\leftarrow$ use Eq. (\ref{eq:proportion_all})     
                \STATE $ \hat{y}_i \leftarrow 1$, with probability $p$ $\forall i: P(1|x_i) = S_k$
            \ENDIF
        \ENDFOR
   \RETURN $\hat{Y}=\{\hat{y}_i |i=1,\dots,N\}$ 
   
\end{algorithmic}
\end{algorithm}

\subsection{Adaptive Cost-Sensitive Learning with Resource Constraints Algorithm}
Our goal is to develop a cost minimizing Adaptive Cost-Sensitive Learning with Resource
Constraints (AdaCSL-WRC) algorithm for decision tree based classifiers. 

The proposed algorithm works as follows. First, in the training phase, a cost-sensitive decision tree based classifier is constructed 
while adjusting the misclassification cost of false positive classifications $c(0,1)$, in each iteration by adding a value of $\epsilon$. In each iteration, the algorithm applies the classifier using the adjusted costs to derive $K$ splits and their respective probability scores $S_k,k=1,...,K$. Then, the classifier dependent threshold $\tau_d$ is calculated using Eq. (\ref{eq:adj_threshold}) and the
instances within the training data set are classified while considering the resource constraint
using Algorithm \ref{alg:classification_based_dynamic_thershold}. To help make the decision whether or not to adjust the misclassification cost
and proceed to the next iteration, the classifier dependent threshold is compared
with an optimal threshold that does not rely on the resource constraint. Once the optimal threshold satisfies the resource constraint, the algorithm
ends at iteration $i$. The generated classifier, at iteration $i=i-1$, is applied to the test data set. We formalize AdaCSL-WRC as Algorithm \ref{alg:AdaCSL-WRC}.

\begin{algorithm}[!h]
            \caption{Adaptive Cost-Sensitive Learning Algorithm for the Resource-Constrained  Classification Problem}
\begin{algorithmic} \label{alg:AdaCSL-WRC}
   \STATE {\bfseries Input:} Training data set $D=(X,Y)$, cost matrix $\mathcal{C}$, decision tree based model $\mathcal{M}$, projected resource constraint on training data set $\mathcal{B}(1)$, step size for cost adjustment $\epsilon$  
   
   \STATE {\bfseries Output:} Learned decision tree based model $\mathcal{M^*}$
   \STATE {\bfseries Initialize:} Cost matrix $\mathcal{C}$, $i$=1
   \REPEAT
   \STATE $\mathcal{M}^i\leftarrow$ $\mathcal{M}\left(D;~\mathcal{C}\right)$ $~~~~~~~~~~~~~~~~~~~~~~~~~~~~~$ //\footnotesize trained model 
   \STATE
   $\mathcal{M}^i$ divide $D$ into subgroups  $D_k, k=1,...,K$ $~~~$//\footnotesize each with the same score $S_k$ 
    \STATE 
    $\tau\leftarrow$ use Eq.(\ref{eq:reduced_threshold})
   \STATE 
$\tau_d\leftarrow$ use Eq.(\ref{eq:adj_threshold}) 
   $~~~~~~~~~~~~~~~~~~~~~~~~~~~~~~~~~~~~~~~~~~~~~~~~~~~~~~~~~~~$
   \STATE
   $\hat{Y} \leftarrow$ use Algorithm \ref{alg:classification_based_dynamic_thershold} $~~~~~~~~~~~~~~~~~~~~~~~~~~~~~~$       //\footnotesize classification of training data set
    \STATE
   $c(1,0)=c(1,0)+\epsilon$
   \STATE
$\mathcal{C}\leftarrow c(0,1);c(1,0)$
   \STATE
   $i = i+1$
   \UNTIL 
   $\tau_d\leqslant\tau$
   \RETURN
   $\mathcal{M}^{*}=\mathcal{M}^{i-1}$
\end{algorithmic}
\end{algorithm}

\section{Experimental Study}
\label{sec:experiment}
Through experiments, we evaluate the performance of a decision tree using the proposed algorithm (AdaCSL-WRC) compared to the common approach of using a cost-sensitive decision tree (denoted CS-DT) and then using its class prediction probabilities in conjunction with a resource-constrained optimization model to allocate label predictions to instances. We compare cost, precision and accuracy indices. For the CS-DT model we use the implementation of \cite{bahnsen2015example} in \href{https://github.com/albahnsen/CostSensitiveClassification/blob/master/costcla/models/cost_ensemble.py}{this GitHub link}.

\subsection{The Data}
The proposed approach was applied to a \href{https://www.kaggle.com/datasets/rodsaldanha/arketing-campaign}{publicly available  data set} obtained from the Kaggle website. 
This data set presents responses of customers to direct marketing campaigns for a retail company. It contains  personal information about customers including, but not limited to, their education, status, and purchase history. In total, the data encompass 2,240 direct marketing campaigns and 23 features after dropping uninformative features, with a single value for all instances or unique values for each instance. The objectives when applying a classification model are to predict who is likely to respond positively to an offer for a product or service (the positive class). The resource constraint represents the limited number of customers that the campaign can target. The frequency of the positive class in the data set is 15\%. The cost of a false positive is the cost of contacting a customer who will not accept the offer and the false negative cost follows from a potential profit loss. We define the misclassification costs arbitrarily to be $c(0,1)=10$ and $c(1,0)=1$, which satisfy $c(0,1)>c(1,0)$. As shown in Section \ref{sec:constraint_line}, a model that produces a lower cost will necessarily be more accurate
 and precise and vice versa when considering a binary classification problem with maximum utilization of a resource constraint. Note that since the classifier dependent value $\tau_d$ is monotonically non-increasing over the iterations in Algorithm  \ref{alg:AdaCSL-WRC} until meeting the stopping criterion, any initial misclassification cost values can be used to achieve the same minimum accuracy and precision, albeit with different convergence times.

\subsection{Experimental Design}
\label{sec:experiment_design}
The experiment was designed such that we could investigate the performance of the proposed algorithm as a function of the two following controlled parameters:
\begin{enumerate}
    \item The number of data set features. This parameter is used to adjust the level of difficulty in classifying tasks. For our experiment, we generated, from the original data set, two reduced data sets that included 75\% and 50\% of the original features, while dropping the features with the highest mutual information values. That is, all the classifiers were evaluated using three data sets: the original data set with all the features, a data set with $75\%$ of the features, and a data set with $50\%$ of the features. 
    \item The value of the resource constraint. We set the value of the resource constraint to [$25\%$,~$75\%$] of the number of campaigns with positive responses. 
\end{enumerate}


For each resource constraint value, we applied a three-fold stratified cross-validation process, repeated 10 times (a total of 30 runs). To evaluate the performance of the models, in each run two subsets were used for training and the remaining subset for testing. The resource constraint for the test data set ranges between $\mathbb{B}(1)\in [28,84]$ and the projected resource constraint for the training data set ranges between $\mathcal{B}(1)\in [56,168]$. Thus, for each model, the classification cost, accuracy and precision were calculated 30 times for each resource constraint value. 
For each model, the hyperparameters were determined based on a grid search. 
The proposed AdaCSL-WRC presented in Algorithm  \ref{alg:AdaCSL-WRC} was applied with a step size for cost adjustment of 1 ($\epsilon = 1$). 
The chosen model generated by the AdaCSL-WRC algorithm is the one applied to the test data set. 



\subsection{Results and Discussion}

\begin{figure}
     \centering
     \begin{subfigure}[b]{0.3\textwidth}
         \centering
         \includegraphics[width=\textwidth]{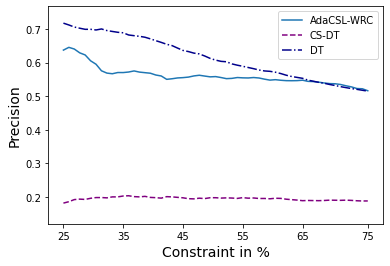}
         \caption{All features}
         \label{fig:marketing_all_prc}
     \end{subfigure}
     \hfill
     \begin{subfigure}[b]{0.3\textwidth}
         \centering
         \includegraphics[width=\textwidth]{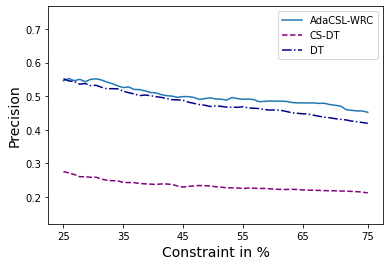}
         \caption{$75\%$}
         \label{fig:marketing_17_prc}
     \end{subfigure}
     \hfill
     \begin{subfigure}[b]{0.3\textwidth}
         \centering
         \includegraphics[width=\textwidth]{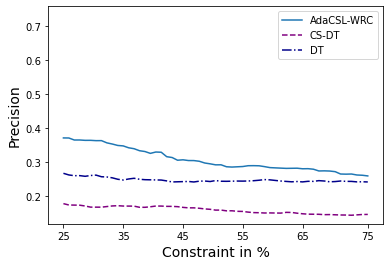}
         \caption{$50\%$}
         \label{fig:marketing_11_prc}
     \end{subfigure}
        
    
    \begin{subfigure}[b]{0.3\textwidth}
         \centering
         \includegraphics[width=\textwidth]{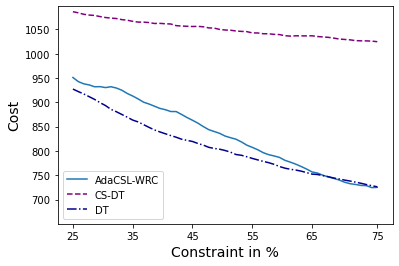}
         \caption{All features}
         \label{fig:marketing_all_cost}
     \end{subfigure}
     \hfill
     \begin{subfigure}[b]{0.3\textwidth}
         \centering
         \includegraphics[width=\textwidth]{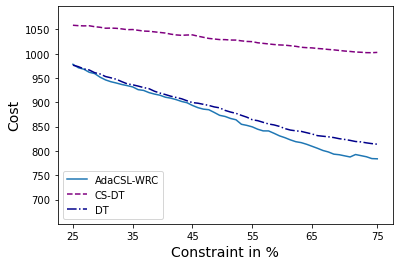}
         \caption{$75\%$}
         \label{fig:marketing_17_cost}
     \end{subfigure}
     \hfill
     \begin{subfigure}[b]{0.3\textwidth}
         \centering
         \includegraphics[width=\textwidth]{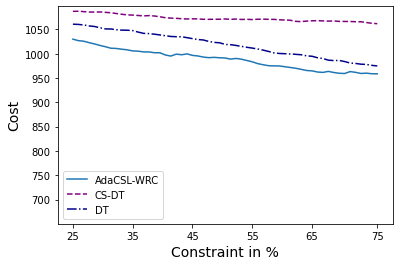}
         \caption{$50\%$}
         \label{fig:marketing_11_cost}
     \end{subfigure}
     
         \begin{subfigure}[b]{0.3\textwidth}
         \centering
         \includegraphics[width=\textwidth]{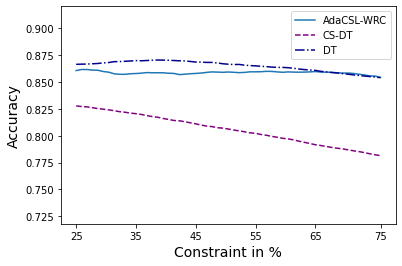}
         \caption{All features}
         \label{fig:marketing_all_acc}
     \end{subfigure}
     \hfill
     \begin{subfigure}[b]{0.3\textwidth}
         \centering
         \includegraphics[width=\textwidth]{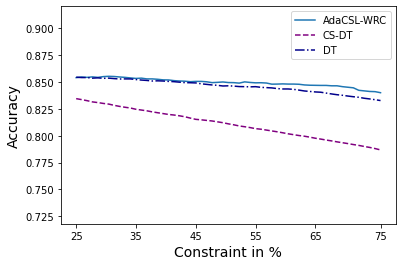}
         \caption{$75\%$}
         \label{fig:marketing_17_acc}
     \end{subfigure}
     \hfill
     \begin{subfigure}[b]{0.3\textwidth}
         \centering
         \includegraphics[width=\textwidth]{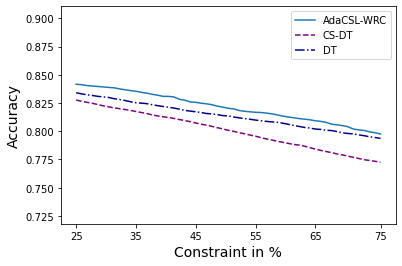}
         \caption{$50\%$}
         \label{fig:marketing_11_acc}
     \end{subfigure}
        \caption{The average precision (graphs a--c), cost (graphs d--f) and accuracy (graphs g--i) over all 30 runs as a function of the resource constraint, for the entire direct marketing campaign data set, the data set with $75\%$ of the features and the data set with $50\%$ of the features.  }
        \label{fig:marketing dataset results}
\end{figure}

\begin{figure}[!t]
	\centerline{\includegraphics[scale=0.5]{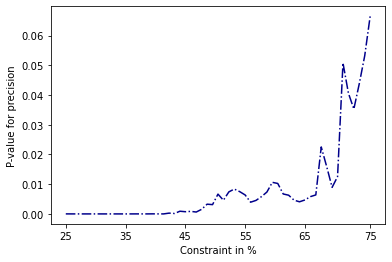}}
	\caption{P-values for comparing the adaptive CS-DT model with a conventional DT model ($50\%$ features case)  for significance.}
	\label{fig:p_val}
\end{figure}

Figure \ref{fig:marketing dataset results} compares the algorithms' performance as a function of the constraint level in terms of the average precision, cost and accuracy for different percentages of the used features. 
When considering all features, which is when the classification problem can be considered to be easier with higher precision and accuracy and lower cost, the preferable model depends on the constraint size. For the range of 25\% to 66\%, the decision tree is the best model, while for the range of 66\% to 75\%, the proposed AdaCSL-WRC is preferable. For the data sets with 75\% and 50\% of the features, the proposed algorithm dominates the other methods over the whole range, and is significantly better (with a p-value<0.05) over the range of 48\% to 75\% for the data set with 75\% of the features and almost over the entire range of [25\% to 74\% for the data set with 50\% of the features, as can be seen in Figure \ref{fig:p_val}.
It is interesting to observe that the proposed AdaCSL-WRC algorithm performs much better in settings where the decision tree model's performance is poor  (e.g., precision below 55\% , reflecting a lift of 367\% or lower, compared to a priori positive instances in the  data set of 15\%).  

To gain insights about the proposed AdaCSL-WRC algorithm compared to the alternative  DT and CS-DT models, we explore the latter's performance in the unconstrained classification problem -- see Table \ref{tbl:base_model_metrics}.
Two interesting insights can be gained from the table. First, the DT model generates lower positive classifications because given that there are fewer features in the data set, the model finds fewer significant patterns resulting in high frequency of classification to the positive class. Therefore, the more difficult it becomes to classify the positive class (minority), the more the proposed AdaCSL-WRC algorithm excels. It yields superior results over the DT model. Second, the CS-DT model generates many more positive classifications than exist in the data set (4.2 to 6.3 times), as result of the high ratio between the misclassification costs, i.e., $c(1,0):c(0,1)=1:10$. Therefore, in a classification problems with a resource constraint below the number of positive instances in the data set (scarce resources), and especially when it is difficult to find patterns that identify a minority class with high probability, our proposed algorithm dominates the alternative CS-DT and DT models.

\begin{table}[!t]
    \caption{CS-DT and DT performance without considering the resource constraint}
    \label{tbl:base_model_metrics}
        \begin{tabular}{ |c|c|c|c|c|c|  }
         \hline
         Model & $\%$ Features & Accuracy & Precision & Cost &  $\%$ Positive  \\
         \hline
         DT
         & $100\%$    & $87\%$ &  $67\%$ & $828\$$  & $40\%$\\
         & $75\%$ & $85\%$ & $53\%$  & $936\$$ & $35\%$\\
         & $50\%$ & $85\%$ & $32\%$  & $1082\$$ & $6\%$ \\
         \hline
         CS-DT
         & $100\%$    & $49\%$ &  $21\%$  & $494\$$ & $421\%$ \\
         & $75\%$ & $45\%$ & $20\%$  & $527\$$ & $445\%$\\
         & $50\%$ & $20\%$ & $15\%$  & $630\$$ & $630\%$\\
         \hline
        \end{tabular}
	\centering
\end{table}

\begin{figure}[!t]
	\centerline{\includegraphics[scale=0.5]{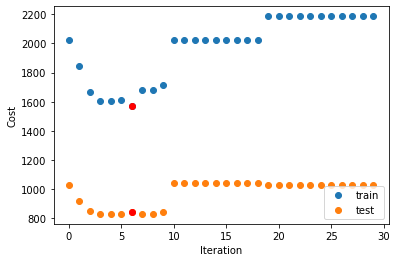}}
	\caption{An example of cost values over iterations of the training and test data sets, received by applying the AdaCSL-WRC algorithm to the data set with all features with a resource constraint of 45\%. The red point represents the final chosen model accepted by the AdaCSL-WRC algorithm.}
	\label{fig:itr_example}
\end{figure}

\begin{figure}[!t]
	\centerline{\includegraphics[scale=0.5]{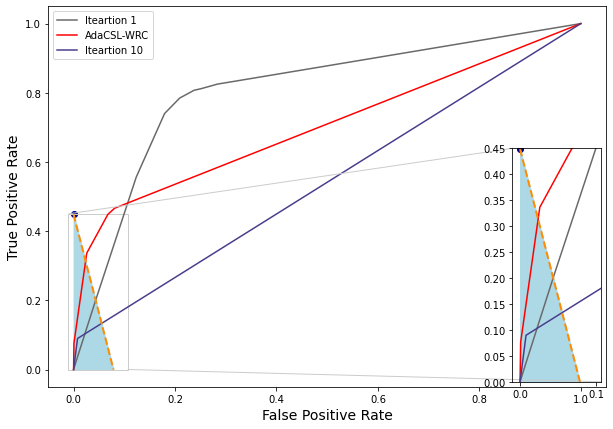}}
	\caption{The ROC curves for the models in the different iterations for the same example as in Figure \ref{fig:itr_example}.}
	\label{fig:roc_example}
\end{figure}

Figures \ref{fig:itr_example} and \ref{fig:roc_example} reflect the adaptive learning of the AdaCSL-WRC algorithm for the data set with all features and a resource constraint of 45\%.
Figure \ref{fig:itr_example} shows the cost values of the training and test data sets over time (iterations). We observe that the cost of the training data set decreases with the iterations, until satisfying the stopping criterion (red dot). The cost of the test data set behaves in a similar way. Since we used three-fold cross-validation, the ratio between the costs of the training and test data sets is similar to the ratio between the sizes of the test and training data sets, i.e., 1:2. Figure \ref{fig:roc_example} presents the ROC curves resulting from the generated models at three different iterations. Specifically, the gray, red and purple ROC curves represent the models generated at iterations 1, 7 (chosen model) and 10, respectively. 
This figure shows that from iterations 1 to 7 (chosen model), the ROC curve intersects the constraint function at a higher point. After reaching the chosen model, however, the ROC curve of the model generated at iteration 10 intersects the constraint function at a lower point. 

\section{Conclusions}
\label{sec:conclusion}
This work develops a novel adaptive cost-sensitive learning approach for solving resource-constrained binary classification problems. The objective is to allocate scarce resources to data instances to minimize misclassification costs and maximize precision and accuracy. 

The proposed approach incorporates a resource constraint on the number of classified instances of classes into the learning process. The algorithm adaptively adjusts the misclassification costs in such a way that the scarce resources are fully utilized and maximum accuracy and precision are achieved. The algorithm was applied, together with a decision tree classifier, to three versions of a direct marketing campaign data set. The resource constraint was modeled as a limited number of customers that a direct marketing campaign can target. The three versions of the data set -- the original data set and data sets with 75\% and 50\% of the features -- represent various difficulty levels classification problems face. Our results show that the decision tree using the proposed AdaCSL-WRC algorithm achieves better cost, accuracy and precision results than other alternative approaches for all resource constraint values, for the data sets representing medium and high levels of classification difficulty. Furthermore, it achieves significantly better results (with p-value<0.05) for most values of the resource constraint. 

Future research directions that we consider are: 1) to apply the proposed approach to other classifiers, and 2) to develop a search method for adjusting the misclassification costs between consecutive iterations.

\bibliographystyle{Styles/elsarticle-num-names} 

\bibliography{Styles/Literature.bib}





\end{document}